\documentclass[letterpaper, 10 pt, conference]{ieeeconf}  % Comment this line out if you need a4paper

\IEEEoverridecommandlockouts                              % This command is only needed if 
\usepackage{amsmath, amssymb}
\usepackage{graphicx}
\usepackage{booktabs}
\usepackage{multirow}
\usepackage{multicol}
\usepackage{pbox}
\usepackage{makecell}
\usepackage{subfigure}
\usepackage[margin=10pt,font=small,labelsep=space]{caption}
\usepackage{wrapfig}
\usepackage{tabularx,ragged2e}
\usepackage{adjustbox}
\DeclareMathOperator{\EX}{\mathbb{E}}% expected value
\DeclareMathSymbol{\shortminus}{\mathbin}{AMSa}{"39}

\usepackage{letltxmacro}
\LetLtxMacro{\oldcite}{\cite}
\renewcommand{\cite}[1]{\mbox{\oldcite{#1}}}

\title{\LARGE \bf Learning from Simulation, Racing in Reality}
\author{Eugenio Chisari$^{1}$, Alexander~Liniger$^{2}$, Alisa~Rupenyan$^{1}$, Luc Van Gool$^{2,3}$, John~Lygeros$^{1}$
\thanks{$^{1}$ Automatic Control Lab, ETH Zurich, Switzerland \newline
 	{\tt{\footnotesize chisarie@student.ethz.ch, \newline \{ralisa,lygeros\}@control.ee.ethz.ch}}}%
\thanks{$^{2}$ Computer Vision Lab, ETH Zurich, Switzerland \newline {\tt \footnotesize \{alex.liniger,vangool\}@vision.ee.ethz.ch}}
\thanks{$^{3}$ PSI, KU Leuven, Belgium}
}

\begin{document}
\maketitle

%===============================================================================

\begin{abstract}
    We present a reinforcement learning-based solution to autonomously race on a miniature race car platform. We show that a policy that is trained purely in simulation using a relatively simple vehicle model, including model randomization, can be successfully transferred to the real robotic setup. We achieve this by using a novel policy output regularization approach and a lifted action space which enables smooth actions but still aggressive race car driving. We show that this regularized policy does outperform the Soft Actor Critic (SAC) baseline method, both in simulation and on the real car, but it is still outperformed by a Model Predictive Controller (MPC) state-of-the-art method. The refinement of the policy with three hours of real-world interaction data allows the reinforcement learning policy to achieve lap times similar to the MPC controller while reducing track constraint violations by 50\%.
\end{abstract}

% \keywords{Reinforcement Learning, Sim-to-Real, Autonomous Driving} 

%===============================================================================

\section{Introduction}\label{intoduction}

Autonomous racing is a quickly growing subfield of autonomous driving where the goal is to drive around a race track as fast as possible. The control policy must be able to perform strategic planning, so that the car is always at the right place on the track with the correct speed. At the same time, the policy has to control the vehicle at the limit of handling, where the behavior of the model is highly nonlinear.
Currently, the most advanced methods use model-based predictive control techniques \cite{liniger2015optimization,rosolia2017autonomous,verschueren2014towards,beal2012model}. Several groups showed that the performance of such methods can be increased by the use of data to improve the prediction model \cite{kabzan2019learning,williams2017information}. Direct learning of a policy that can control an autonomous race car is rarely studied in the literature, an exception being \cite{pan2020imitation}, where a pixel-to-action policy is trained through imitation learning on a 1:5 scale car and \cite{fuchs2020super} where a policy is learned in a race car simulation game.

Compared to model-based approaches, Reinforcement Learning (RL) does not require an accurate vehicle model or a well tuned reward function. Instead, the RL agent learns to race by interacting with the environment using high level rewards. This has been demonstrated on real robots for several robotics applications \cite{peng2018sim,akkaya2019solving,haarnoja2018learning,hwangbo2019learning,waslander2005multi,hwangbo2017control,bou2010controller,kendall2019learning,baltes1999path,kamran2019learning}, in areas like robot manipulation, quadcopters, and wheeled robots.
When a model of the system is available, the RL agent can be trained in simulation with the purpose to transfer the learned policy on a robotic system \cite{tobin2017domain} using the \emph{Sim-to-Real} approach. However, the RL agent learns to interact with a limited representation of the environment and is prone to over-fitting with respect to the simulation. Tools such as domain or model randomization \cite{tobin2017domain,akkaya2019solving,loquercio2019deep} often drastically help the sim-to-real transfer, by randomizing the observation and model of the simulator. In \cite{bewley2019learning,osinski2019simulation,amini2020learning} it was shown that sim-to-real works for autonomous driving, however, their main challenge was the vision domain gap, where in our autonomous racing case the domain gap in the vehicle model is the main challenge.

Regularization is an essential ingredient guiding the learning process, shaping the corresponding policy, and incorporating domain knowledge in the RL problem. In RL, the entropy maximization regularizer \cite{willims1991} is commonly used to encourage exploration. Recently, the effect of other regularization strategies popular in deep learning has been studied in \cite{liu2019regularization}. They showed that using regularizers such as weight clipping, dropout, or BatchNorm can be beneficial, but had the best results when using a \(L_2\) regularization on the policy weights, even outperforming the classical entropy regularizer.
In \cite{wang2018reinforcement, zhu2020safe}, the reward function was augmented using hand crafted features in order to achieve smooth driving behavior.

This paper shows how we can train an autonomous racing policy using sim-to-real and apply it to control a miniature racing car. Directly using an RL method to learn this task is challenging because the system has continuous states and actions, requires a refresh rate of 100 Hz, has highly uncertain dynamics, and finally, the boundaries of the race track are walls, making exploration difficult. We show that a policy learned with an off-the-shelf actor critic method can be generalized to the real system, even if the agent is only trained in simulation with a simple vehicle model. We achieve this by using a well-suited state and action space in combination with model randomization, and by incorporating a novel policy output regularization technique, motivated by control theory. This regularizer, combined with input-lifting techniques where the actions are the changes in the physical inputs, significantly improves the sim-to-real transfer. We compare the proposed regularizer with established techniques such as reward shaping or \(L_2\), and show that our method performs better in simulation and on the real robotic platform. Finally, we show that the policy learned in simulation can be refined by interacting with the real environment, achieving a performance comparable to that of a state of the art predictive controller in terms of lap time, while achieving 50\% less constraint violations.
%===============================================================================

\section{Autonomous Racing}\label{sec:methodology}

We now introduce the experimental setup, and define the related Markov Decision Process (MDP). The experimental platform consists of a miniature 1:43 remote controlled Kyosho dNaNo car. The car is controlled via Bluetooth by a remote PC and the position, heading and velocities are estimated by an overhead infra-red camera system at 100Hz. The car reaches speeds of above 3~m/s, which correspond to an upscaled speed of 465~km/h. The car drives around a 18~m long race track with 13 turns, see \cite{LinigerPhD} for more details. Due to the small size and high relative speeds, the car is challenging to control because its behavior is highly non-linear, making it difficult to model and predict the dynamics accurately. Previous research showed that considering this uncertainty can drastically help \cite{carrau2016,hewing2019}.

\subsection{State Space and Action Space}

We are interested in finding an action space where smoothness can be enforced in the actions, and a state space where discontinuities in the (action) value function and in the policy can be avoided.
We therefore represent the state in a curvilinear coordinate system \cite{vazquez2020optimization} (Frenet frame), which transforms the position and heading relative to a reference path, in our case the center line of the race track. More specifically, we have that \( p \) is the progress along the track, \( n \) is the deviation from the reference path, and \( \mu \) the heading relative to the path. Following standard vehicle dynamics, we also consider the longitudinal and lateral velocity in the vehicle's frame, \( v_x \) and \( v_y \) respectively, and the yaw rate \(\omega\) as states. The physical control inputs to our car are the duty cycle to the rear wheel drive motor \(d\), and the steering angle \( \delta\). Since jerky inputs could damage the actuators and harm the performance, we lift our inputs and add \( d \) and \( \delta \) to the states and consider the rate of change of the physical inputs \( d_{rate} \) and \( \delta_{rate} \) as our actions. In summary, our state is given by \(s = [p, n, \mu, v_x, v_y, \omega, d, \delta]\), and the actions by \(a = [d_{rate}, \delta_{rate}]\).
This approach allows to efficiently penalize jerky inputs to our autonomous car (as we will discuss in Section \ref{sec:regularization}). We did not experience a noticeable slow down in learning because of this addition.
Finally, the actions (input rates) as well as the physical inputs are both constrained. In the case of our miniature race cars we have the following bounds, \(d \in [-0.2, 1.0], \delta \in [-0.35, 0.35], d_{rate} \in [-17.5, 17.5],\) and \(\delta_{rate} \in [-3.5, 3.5]\). We implement the constraints on the input rates using a $\tanh$ output layer in our policy and the physical input constraints by clipping. Note that we assume that we can observe the state, which is reasonable given the overhead camera system, but even for full-size race cars an autonomous driving stack similar to \cite{kabzan2019amz} could be used to achieve the same.

\subsection{Reward Function}

The goal of a race car is to drive around the track as quickly as possible, while not violating the track limits. In this work we propose to formulate this objective using a dense reward function that penalizes constraint violations. Therefore, let us first define the constraints used. The first is a track constraint which is triggered if the deviation to the center line is larger than half the track width \( |n_{t+1}| \leq w_{\text{tr}}/2 \). Note that in practice we added a safety margin to the actual track width, since in our experimental setup the boundaries are physical walls, and touching them either leads to an accident or drastically slows down the car. The second constraint we introduce is added to facilitate the sim-to-real transfer. When driving at the limit of handling, combined slip of the tires is important, but it is not modeled in our simulator. Thus, we include a constraint in our reward that penalizes excessive combined slip. We model this using the following tire ellipse constraint \( F_{y, R}^2 + {(p_\text{long} F_x)}^2 \leq {(p_\text{ellipse} D_\text{R})}^2 \), where \( p_\text{long}  = 0.9\) and \( p_\text{ellipse} = 0.95\) are tyre specific parameters \cite{kabzan2019amz}.
To model the lap time minimization objective, we use an incremental progress reward \(p_{t+1} - p_t\), where the agent is rewarded to drive as far as possible with respect to the center line. This reward is popular in autonomous racing \cite{liniger2015optimization}, and it is a good dense approximation of the sparse minimum lap time reward. Thus our reward function is defined as
\[
    \begin{aligned}
        r_t =
        \begin{cases}
            -c            & \text{if constraints violated} \\
            p_{t+1} - p_t & \text{otherwise }
        \end{cases}
    \end{aligned} \,,
\]
where $c = 0.01$ and $p_{t+1} - p_t$ is in the range between 0 and 0.05.
The proposed reward penalizes deviations from the track and driving in regions of the state space where our simulator is not precise, while it rewards fast driving.

\section{Sim-to-Real Method}

In the following section we explain the necessary steps to perform the sim-to-real transfer for our autonomous racing task. We also introduce a novel policy regularization approach to facilitate the sim-to-real transfer.

\subsection{RL Setup}

We formulate our problem as an episodic RL problem, where an episode is either terminated if the car violates the track constraints for more than double the track width or if it lasts for 600 time steps (equivalent to 6\,s of driving). To improve exploration of the state space we start each episode at a random initial state and use relative short episodes, following \cite{sutton2018reinforcement}. We use Soft Actor Critic (SAC) \cite{haarnoja2018soft} due to its superior performance compared to other state of the art RL methods on our setup, and due to its off-policy capabilities which we will exploit in Section \ref{sec:refinement}. Our implementation of SAC is based on Stable Baselines~\cite{stable-baselines}, and we implement our simulator, explained in Section \ref{sec:mod_rand}, as an OpenAI Gym environment \cite{gym}. The control policy and (action) value functions, are implemented as fully connected neural networks with two hidden layers and 256 neurons per layer. We use a learning rate of \(3 \cdot 10^{-4}\), a batch size of 512, a replay buffer of length \(10^{6}\), and a discount factor of 0.99. We run SAC at every time step of the simulation and train for $1.8 \cdot 10^7$ time steps (50~hours of simulated driving). We run the training on a single desktop computer, where we only use the CPU (Intel i9-9900K), which takes about 36~hours.

\subsection{Simulator and Model Randomization}\label{sec:mod_rand}

Our simulator is based on the dynamic bicycle model proposed in \cite{liniger2015optimization}, which uses Pacejka tire models \cite{pacejka1992magic} to model the interaction with the ground.
This model is simpler than full vehicle dynamic models, which makes it computationally more tractable. However, it is generally over-optimistic and does not capture the uncertainty of the real cars.

Therefore, we use model randomization to train a policy that can transfer to the real system, by reducing the reality gap. We propose to use a multiplicative uncertainty model of the form $\dot{s} = f(s, a) (1 + \varepsilon)$, where \( f(s, a) \) is the nominal dynamic bicycle model and \( \varepsilon \) the uncertainty. The advantage of a multiplicative uncertainty is that in the case of rapid changes in the system, which are difficult to model, the effect of the uncertainty is even increased. Since the position and orientation are only the integration of the velocities, we consider only uncertainty on the velocity states. We model this uncertainty as a uniform distribution, where the bounds are determined from real driving data recorded using a benchmark MPC. Given the state-action sequence the one-step model error can be computed using an Euler forward integrator. Based on these errors we chose the noise distribution to approximately include 80\% of all recorded errors. When using uncertainty bounds that include all recorded errors or even overestimate the noise level, the resulting policies are over conservative. The resulting uniform distribution we use in training are \(\varepsilon_{v_x} = \mathcal{U}(-1.5,1.5)\), \(\varepsilon_{v_y} = \mathcal{U}(-2.5,2.5) \), and \(\varepsilon_{\omega} = \mathcal{U}(-2.0,2.0)\).

\begin{figure}[h]
    \centering
    \subfigure{\includegraphics[width=0.225\textwidth]{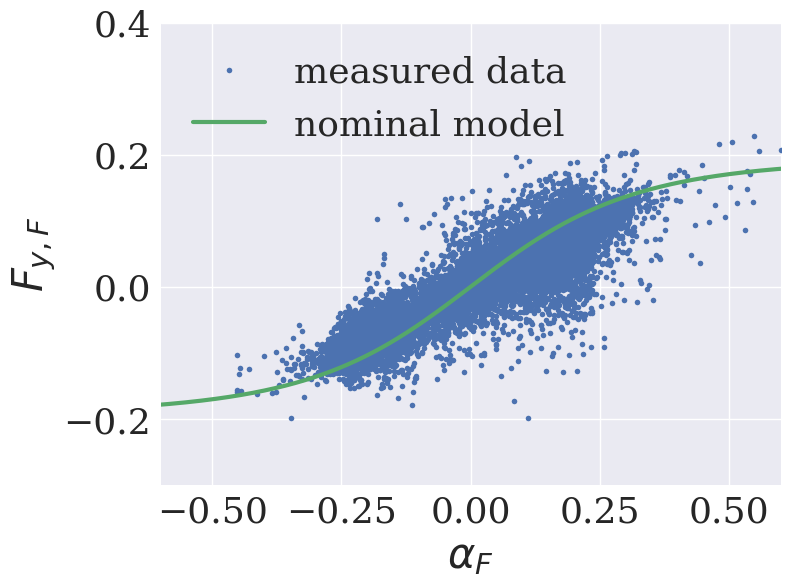}}
    \subfigure{\includegraphics[width=0.225\textwidth]{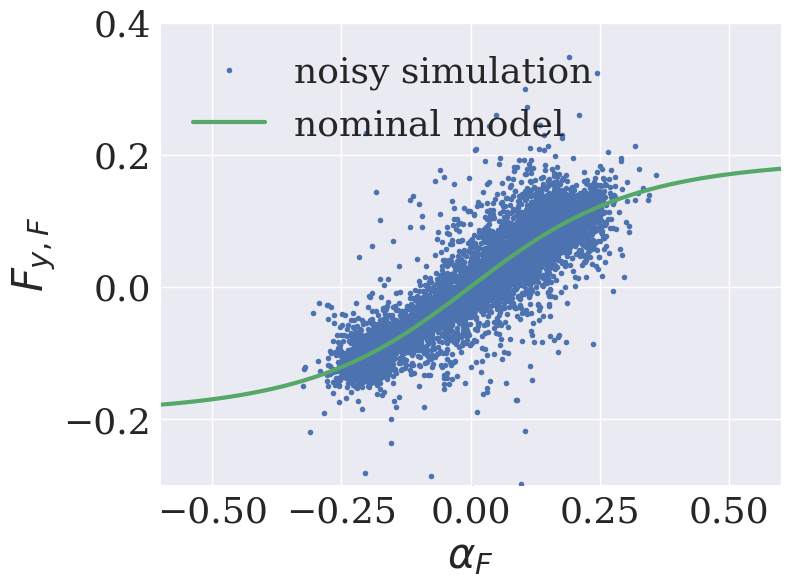}}
    \caption{Estimated tire forces vs Pacejka tyre model, real measurements (left) and noisy simulation (right), both using our trained sim-to-real policy.}\label{fig:tyre_model}
\end{figure}

To qualitatively demonstrate the effect of our model randomization technique, we estimate the tire forces from state-action sequences, using a model inversion technique, and compare these force estimates obtained from experimental and simulation sequences. Fig.~\ref{fig:tyre_model} on the left shows the front wheel force estimates versus the slip angle when driving the real car as well as the nominal Pacejka tire model used in our simulator for reference. It is clearly visible that the real platform has a considerable degree of uncertainty, which makes model randomization a simple but necessary tool to bridge the sim-to-real gap. The force estimate of our model randomization approach is shown in Fig.~\ref{fig:tyre_model} on the right. We see that the uncertainty we use in our simulator resembles the one appearing in the real recorded data.

\subsection{Policy Regularization}\label{sec:regularization}

There are two main motivations for regularizing action sequences. First, excessive and jerky inputs can potentially damage actuators, and in autonomous racing, the fast changing inputs can also introduce drag that degrades performance. The second reason is related to the transfer from simulation to reality. Excitation of a system with high-frequency inputs can result in behaviors that are difficult to model. In our experimental setup this is for example the load transfer, which is not considered in the simulation model. To mitigate these effects, it is common in model-based control approaches to consider an input cost that penalizes the control inputs. In RL-based approaches, however, such regularizers are less common, but have recently attracted some interest \cite{Cheng2019}.

\subsubsection{Reward and Policy Weight Regularization}
Let us first introduce two standard approaches to regularize the inputs. First, we can augment the reward function, and add a quadratic cost on the applied actions, $r_t^\text{rar} = r_t - a^T \tilde{M} a$. The matrix $\tilde{M}$ is a hyper-parameter and we call this method \emph{reward action regularization}. The second approach is to use standard regularizers from deep learning, the most popular one being the \(L_2\) penalty on the policy weights, which we call from now on \emph{policy weight regularization}.

Both approaches have disadvantages. The \emph{reward action regularization} approach results in slow learning, because the RL agent has to learn if a low reward is caused by non promising states or too excessive inputs. The \emph{policy weight regularization} approach is tailored to reduce over-fitting but it is not effective at achieving our goal of smooth actions.

\subsubsection{Policy Output Regularization}

\begin{table*}[ht]
    \centering
    % \vspace{4mm}
    % \renewcommand\baselinestretch{1.0}
    \begin{tabular*}{0.95\textwidth}{l c c c c c c c c c c c c c c c}

        \toprule
        & \multicolumn{1}{c}{No Reg.} & \multicolumn{4}{c}{Output Reg.} & & \multicolumn{4}{c}{Reward Reg.} & & \multicolumn{4}{c}{Weight Reg.}\\
        \cmidrule{3-6}  \cmidrule{8-11} \cmidrule{13-16}
        Reg. value&  &  1 & 5 & 10 & 50 & & 0.1 & 0.2 & 1 & 2 & & 1e-7 & 1e-5 & 1e-4 & 0.001\\
        \midrule
        \textbf{Mountain Car Cont.}\\
        avg. reward &  70   & 100  & 100  & 100   & -  & & 100 & -  & - & -  & & 100 & 100 & 100 & 100 \\
        avg. $\dot{a}_\text{rms}$  &  5.2  & 2.5 & 4.2 & 1.8 & - & & 3.6 & - & - & - & & 1.5 & 1.7 & 1.7 & 1.6 \\
        \midrule
        \textbf{Pendulum}\\
        avg. reward &                -147   & -164  & -152  & -166  &  -152 & & -148 & -151 & -146 & -153 & & -155 & -148 & -147 & -150 \\
        avg. $\dot{a}_\text{rms}$  &  31.6  & 11.8   & 9.7    & 15.4   &  7.5  & &  10.7 & 10.7  &  10 &   6.7 & &  13.6 &  11.2 &  11.0 &   9.9 \\
        \midrule
        \textbf{PyBullet Reacher}\\
        avg. reward &      14.0   & 15.4  & 14.7  & 13.7  &  16.0 & & 12.5 & 14.7 & 14.4 & - & & 12.4 & 12.5 & 14.5 & 13.1\\
        avg. $\dot{a}_\text{rms}$  &  70.0 & 31.8& 14.7   & 6.4  &  1.6 & &  71.6 & 37.1&  3.2 & - & &  83.2 & 76.1 &  51.3 &  76.5\\
        \bottomrule
    \end{tabular*}
    \caption{Benchmark performance for different regularizers. }\label{tab:RegInfluence}
\end{table*}

Given these insights we propose a novel regularization approach, \emph{policy output regularization}, which directly regularizes the policy output without slowing down learning by using gradient information in the policy update step. More precisely, we try to find a policy that minimizes the following objective,
\begin{align*}
    \pi_M^*(s_0) & = \underset{\pi}{\mathrm{argmax}} \ -\underset{}{\EX} [\pi(s_0)]^T M \underset{}{\EX} [\pi(s_0)] \\
                 & + \underset{\tau \sim \pi}{\EX} \left[ \sum_{t=0}^{\infty}
        \gamma^t \bigg( R(s_t, a_t, s_{t+1}) + \alpha H\left(\pi(s_t)\right) \bigg)\right] \,,
\end{align*}
where \(M\) is a positive semidefinite matrix and \(\alpha H\left(\pi(s_t)\right)\) is the standard SAC entropy term. Compared to SAC, we only add the first quadratic term that penalizes high values of the actions applied on the first time step. Through the choice of \(M\), the regularization can be tuned independently for each action ($d_{rate}$ and $\delta_{rate}$ in our case).
We can derive the regularized action value function as
\begin{align}
    Q_M^\pi & (s_0, a_0) = - a_0^T M a_0 \nonumber                                \\
    +       & \underset{\tau \sim \pi}{\EX} \left[ \sum_{t=0}^{\infty} \gamma^t
        R(s_t, a_t, s_{t+1}) + \alpha \sum_{t=1}^{\infty} \gamma^t
        H\left(\pi(s_t)\right) \right]
    \\
    Q_M^\pi & (s_0, a_0) = - a_0^T M a_0 + Q^\pi(s_0, a_0) \label{eq:Qm_relation}
    ,\end{align}
where \( Q^\pi \) is the standard SAC action value function.
Analogously to the standard SAC policy gradient update, the update of the regularized policy is
\begin{align}
     & \nabla_{\theta} \frac{1}{|B|} \sum_{s_t \in B} - \left[Q_{M, \phi}\left(s_t, \pi_{\theta}(s_t)\right)
        + \alpha H\left(\pi_\theta(s_t)\right) \right],
    \label{eq:reg_loss}
\end{align}
with $B$ the sampled batch.
Equation~\eqref{eq:Qm_relation} shows a tight relation between the standard action value function \( Q^\pi \) and the regularized version \( Q_M^\pi \). Given this relation, the policy loss~\eqref{eq:reg_loss} can be rewritten as
\begin{align}
    \nabla_{\theta} \frac{1}{|B|} \sum_{s_t \in B} - & \left[Q_{\phi}\left(s_t, \pi_{\theta}(s_t)\right)
        + \alpha H\left(\pi_\theta(s_t)\right) \right] \nonumber                                                                      \\
    +                                                & \underset{}{\EX} [\pi_{\theta}(s_t)]^T M \underset{}{\EX} [\pi_{\theta}(s_t)].
\end{align}
Therefore, to train a regularized policy, we can use the standard SAC algorithm, see Algorithm 1 in \cite{haarnoja2018soft}, to learn \( Q^\pi \) and add a quadratic term in the policy gradient step.

This regularizer can be used without input lifting. For our system and other robotic systems, however, the combination of considering input rates as actions and regularizing the policy output is preferable, since this approach does not penalize physical inputs but the rate at which they change. Therefore, a constant steering or the use of a high torque when holding a heavy object with a robot arm is not penalized, but a jerky steering or torque is. We study the discussed regularizers in more details on common RL benchmarks in Section \ref{app:reg} and demonstrate that the policy output regularizer has a positive effect both on the smoothness of the actions and on the reward, and outperforms the other regularizers in the more complex tasks.

\section{Sim-to-real Results}

We first present results how our proposed \emph{policy output} regularization, compares to the other two regularizers in a simulation environment. To get a better insight we study the regularizers both on our autonomous racing problem and on three benchmark tasks. After our simulation results we present the experimental results, for the autonomous racing problem, showing the benefit of our regularizer in the sim-to-real transfer.

\subsection{Policy Regularization for Benchmark Task} \label{app:reg}
% \vspace{-1mm}
To highlight the importance of policy/action regularization we first studied the presented methods, \emph{reward action}, \emph{policy weight} and our proposed \emph{policy output} regularization on three RL benchmark tasks. We tested two OpenAI gym problems \cite{gym}, \emph{MountainCarContinuous-v0} and \emph{Pendulum-v0}, which are both simple RL problems. We followed the suggested RL-zoo \cite{rl-zoo} hyper-parameters and trained for $6 \cdot 10^4$ steps. The third benchmark is PyBullet Reacher \cite{pybullet}, where we used the default SAC settings and trained the policy for $5 \cdot 10^5$ steps. For all three examples we tested vanilla SAC, and 4 different levels of regularization for the three tested regularizers, see Table \ref{tab:RegInfluence} for the used regularization values\footnote{Diagonal matrices with these values when necessary.}. After training, we evaluated 100 rollouts, with resulting average reward and mean RMS of the actions derivative reported in Table \ref{eq:reg_loss}. Note that when the RL agent did not learn the task we report $-$ instead. The RMS of the actions derivative can be understood as a smoothness score, where lower values mean smoother actions. It is highly relevant for this work, as we will show that in the race car application smooth inputs facilitate the sim-to-real transfer. Note that compared to the race car setup we did not lift the actions, in order to leave the benchmarks unchanged.

In all three benchmarks, regularization helps not only to get smoother actions but also to improve the reward achieved by the policy. For the first two tasks \emph{policy weight regularization} achieved the best results closely followed by our \emph{policy output regularization}. Similar to our race car results (see section \ref{sec:reg_car}, Table \ref{table:inputs}) \emph{reward action regularization} often results in the RL agent not learning: there seems to be a very small range of hyper-parameters where this approach works, which makes it unpractical.
In the last and hardest task, Reacher, we can clearly see that our propose \emph{policy output regularization} achieved the best results. The main difference is that \emph{policy output regularization} can achieve smooth actions without negatively impacting the learning. This stands in contrast to \emph{reward action regularization} which is not designed to enforce smooth actions and failed in this task. Even though these results do not evaluate the effect on sim-to-real, we can see that our regularizer helps in all tasks to achieve better policies which act in a smooth fashion.

\subsection{Policy Regularization for Simulated Autonomous Racing}
\label{sec:reg_car}

\begin{table}[ht]
    \centering

    \renewcommand\baselinestretch{1.0}\selectfont
    \begin{tabularx}{0.47\textwidth}{m{2cm} c c c c c} % 6 columns
        \toprule
         & SAC
        % & {\thead{SAC + Reward \\ Action Reg.}}
        % & {\thead{SAC + Policy \\ Weights Reg.}}
        % & {\thead{SAC + Policy \\ Output Reg.}}
         & {\thead{Reward  \\ Reg.}}
         & {\thead{Weights \\ Reg.}}
         & {\thead{Output  \\ Reg.}}
         & MPC
        \\ \midrule
        \small{Lap Time \([s] \) }
        %\thead{Lap Time \([s]\)}
         & 10.0
         & 11.3
         & 9.8
         & 9.8
         & 8.7
        \\
        {\small{Constraint Viol. \([s {\times} 10^{\shortminus2}] \) }}
         & 3.6
         & 15.6
         & 9.0
         & 1.7
         & 8.6
        \\
        \( d_\text{rms} [10^{\shortminus1}] \)
         & 5.8
         & 4.5
         & 5.9
         & 5.5
         & 7.6
        \\
        \( \delta_\text{rms} [rad {\times} 10^{\shortminus1}]\)
         & 2.0
         & 2.0
         & 2.0
         & 2.0
         & 2.3
        \\
        \( \dot{d}_\text{rms} [s^{\shortminus1} {\times} 10^0]\)
         & 6.8
         & 2.8
         & 6.3
         & 3.6
         & 5.5
        \\
        \(\dot{\delta}_\text{rms} [\frac{rad}{s}]\)
         & 2.3
         & 1.9
         & 1.7
         & 1.8
         & 1.7
        \\
        \( \ddot{d}_\text{rms} [s^{\shortminus2} {\times} 10^1 ]\)
         & 66.6
         & 12.6
         & 55.0
         & 22.1
         & 30.8
        \\
        \(\ddot{\delta}_\text{rms} [\frac{rad^2}{s^2} {\times} 10^1]\)
         & 14.8
         & 12.6
         & 9.7
         & 10.5
         & 4.9

        \\ \bottomrule \\
    \end{tabularx}
    \vspace{-0.5cm}
    \caption{Comparison of performance and RMS action smoothness metrics for different policies}\label{table:inputs}
\end{table}

\begin{figure}[h]
    \centering
    \includegraphics[width=0.4\textwidth]{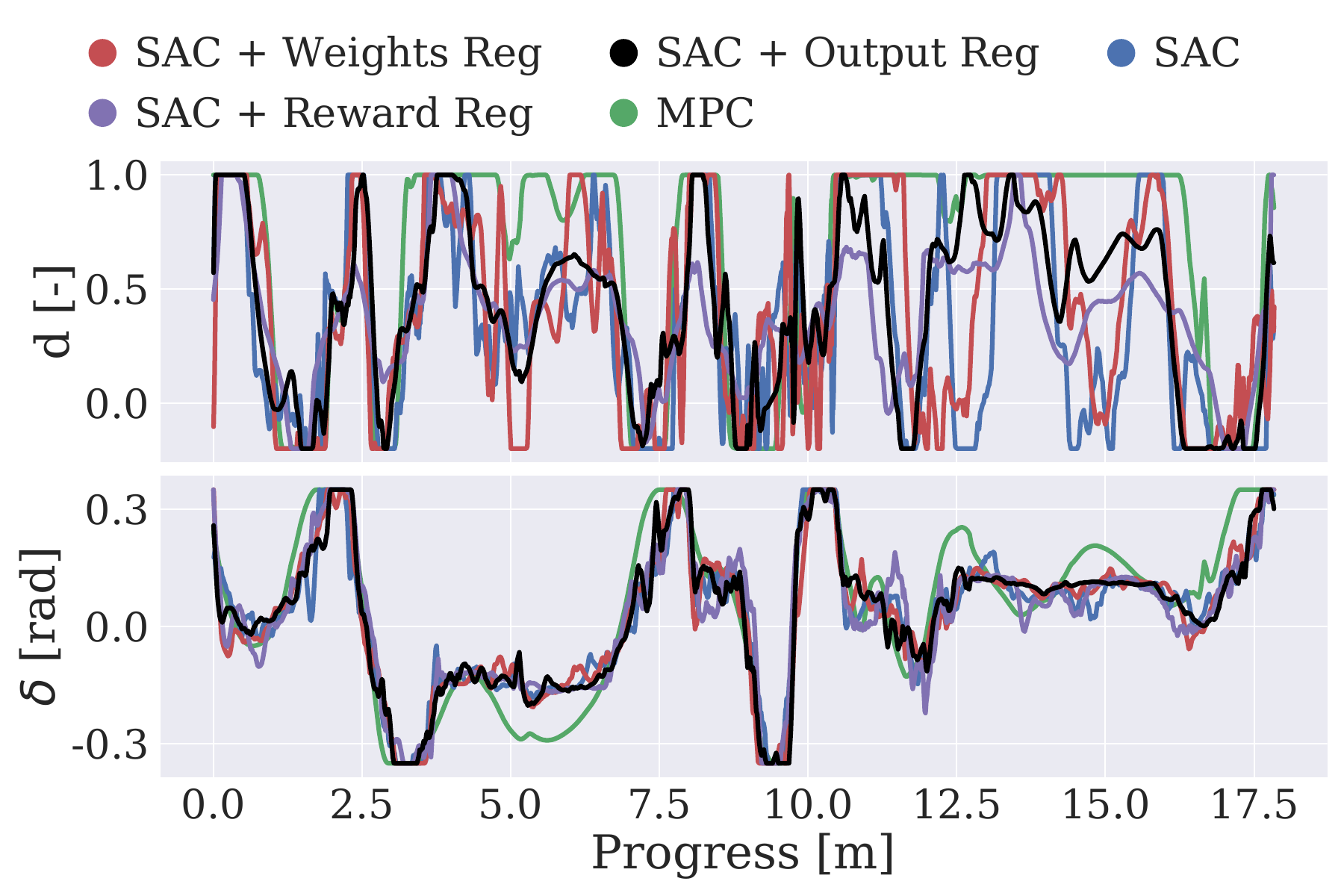}\\
    \caption{Comparison of \(d \) and \(\delta\) over one lap.}\label{fig:inputs_comparison}
\end{figure}

To compare the different regularization methods we train four different policies, (i) vanilla SAC, (ii) SAC with reward action regularization \eqref{eq:reg_loss} ($\tilde{M} = \text{diag}(0.005,0.001)$), (iii) SAC with policy weight regularization (\(L_2\) penalty \(10^{-4}\)),  and (iv) our proposed method SAC with policy output regularization ($M = \text{diag}(50,10)$). Finally we compare the results with the state of the art MPC method \cite{liniger2015optimization}. All five methods are evaluated in the same simulation environment \emph{with} model randomization present.
In Table \ref{table:inputs} we can see the results, where we show our two main driving objectives, lap time and constraint violation. The latter shows how many seconds the car is closer than 2\;cm to the wall per lap. Since the goal of this study is also to evaluate the policy regularizers, we report the Root Mean Square (RMS) of the physical inputs ($d, \delta$), and the RMS of the first ($\dot{d}, \dot\delta$) and second derivative ($\ddot{d}, \ddot\delta$), similar to Table \ref{tab:RegInfluence}.
We can see that the MPC has the lowest lap time, but the third highest number of constraint violations. The four RL based methods are significantly slower, with method (ii) having the highest lap time, while also learning very slowly. The number of constraint violation of the four methods spans a large range, again with (ii) showing the worst performance. However, method (i) and especially our proposed method (iv) show significantly less constraint violations compared to the MPC controller. Method (iii) stands in between with the same lap time as (iv) but higher constraint violations.
When comparing the smoothness metrics, we can mainly investigate the second order derivative, where we can see that vanilla SAC (i) is the worst performing, and the RL methods with additional regularization clearly generate smoother inputs. This is also shown in Fig.~\ref{fig:inputs_comparison}, where SAC has clearly the most jittery actions. Method (ii) achieves a smooth duty cycle by driving slow, but has only a slightly smoother steering than (i). Method (iii) has a similar pattern as the MPC which is desired, but the smoothness scores are twice as high, which is problematic for the sim-to-real transfer. Finally, method (iv) has a smooth duty cycle, but aggressive steering. Even though this is a strategy different than the MPC, it allows for a successful sim-to-real transfer. We believe that the RL agent has learned that a fast, aggressive steering that corrects errors quickly, coupled with a relatively smooth duty cycle input works well. The input trajectories of our proposed method (iv) in Fig.~\ref{fig:inputs_comparison} look visibly better than the other RL based methods, especially the $d$ action is smoother and has significantly less jitter. The steering input has a shape similar to that of the other RL agents, but again with reduced jitter.

\subsection{Sim-to-Real Experimental Results}

\begin{figure}[h]
    \centering
    \includegraphics[width=.375\textwidth]{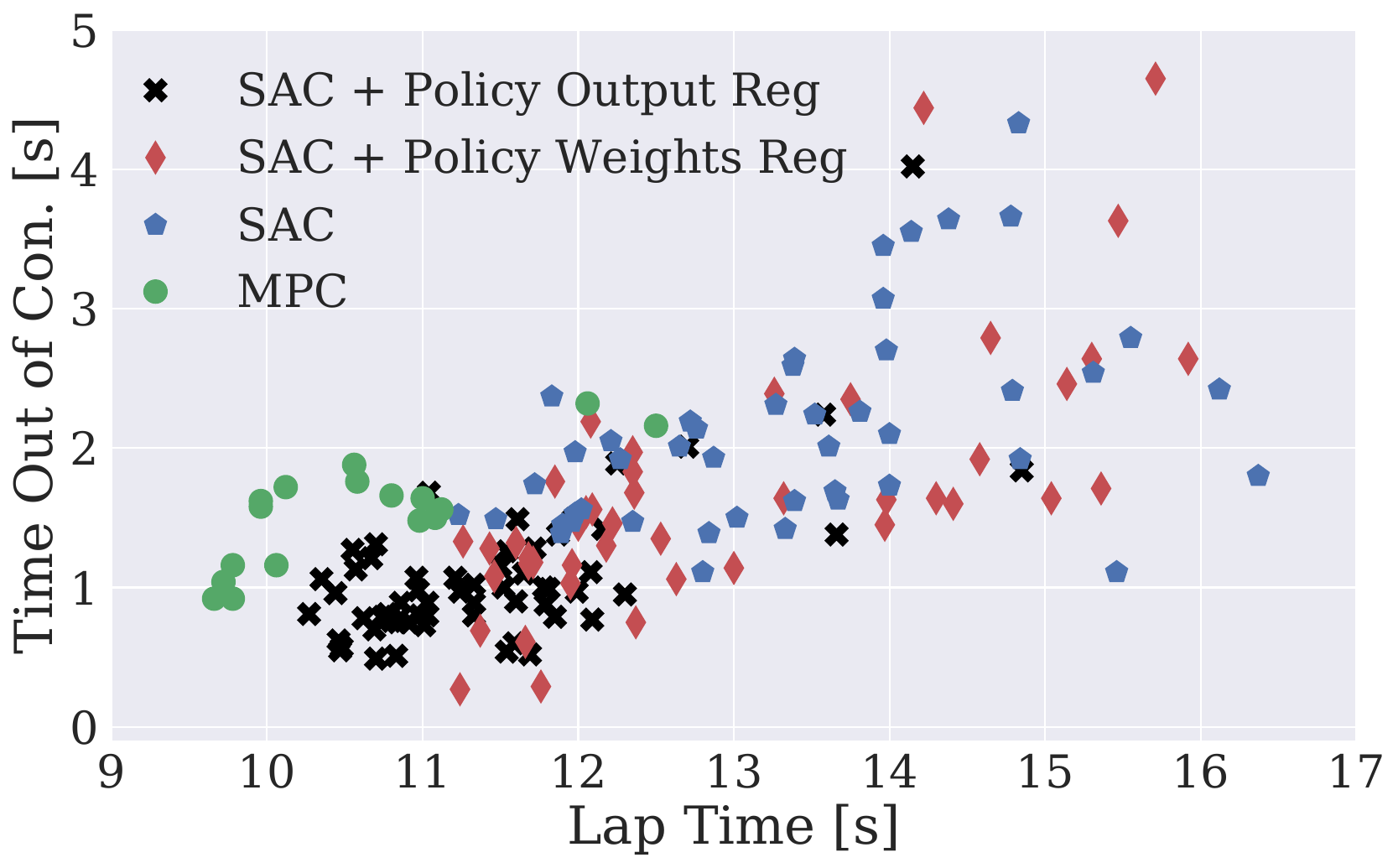}\\
    \caption{Performance of different policies trained in simulation.}\label{fig:performance_sim}
\end{figure}

We now show that model randomization and the proposed \emph{policy output regularization} facilitate sim-to-real transfer, by testing the trained polices discussed in the previous section on our experimental platform. The performance on the real platform of the policies trained in simulation is shown in Fig.~\ref{fig:performance_sim}. We compare the mean lap time and time outside of the track constraints achieved by different learned policies, and by a benchmark MPC controller~\cite{liniger2017real}.
The benchmark MPC is a hierarchical method consisting of a recursive roadmap motion planner with a terminal viability constraint and a 0.32s look ahead, combined with nonlinear MPC trajectory tracker. Note that for both the MPC and the RL policy the time delays are compensated using a Kalman filter approach \cite{LinigerPhD}.
Lower lap times and constraint violations correspond to better performance. Data points with large lap time are an indication of collisions with the track walls. All policies in the figure are trained with model randomization. Without model randomization (not shown), the policy is not able to generalize, and the car is not able to complete a lap. We also exclude the experimental results of the reward action regularization approach due to its bad performance. Using vanilla SAC\footnote{What we call SAC in our results can be considered ``standard" sim-to-real with model randomization} results in a 28\% higher mean lap time, and 40\% higher constraint violations, compared to the MPC benchmark policy. Policy weight regularization of the policy results in 24\% slower lap times and 17\% more constraint violations compared to the MPC, showing that policy regularization can help. The policy that performs best is the one trained with policy output regularization. It achieves a performance comparable to that of the MPC controller we use as benchmark, with a 30\% improvement in constraint violations, and only 8\% slower lap times on average. While the performance achieved with policy output regularization is not far from the performance of the MPC controller, it cannot match the peak performance of the MPC.
\vspace{-3 mm}
\begin{table}[h]
    \centering
    % \caption{Sim-to-Real performance}
    % \label{table:Sim2Real}
    \begin{tabular}{lcc} % 3 left aligned columns
        \toprule
        {}                       & {\thead{Average                 \\ Lap Time [s]}}
                                 & {\thead{Average Time Out        \\ of Constraints [s]}} \\
        \midrule
        SAC                      & 13.53                    & 2.15 \\
        SAC + Policy Weights Reg & 13.18                    & 1.79 \\
        SAC + Policy Output Reg  & 11.43                    & 1.07 \\
        MPC                      & 10.57                    & 1.53 \\
        \bottomrule
    \end{tabular}
\end{table}
% \vspace{-4mm}
\vspace{-5 mm}
\section{Policy Refinement on the Car}\label{sec:refinement}

\begin{figure}[h]
    \centering
    \includegraphics[width=0.4\textwidth]{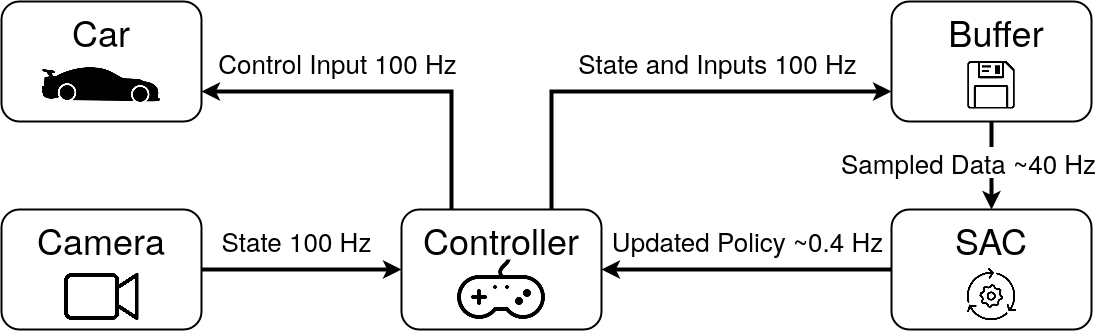}\\
    \caption{Online reinforcement learning architecture}\label{fig:online_architecture}
\end{figure}

Given that we can transfer a policy learned in simulation to the real physical car, RL can now be used to refine this policy by using the real physical system, without the issues related to learning a policy from scratch. This allows to overcome the limitations of the simulator and train the system in a fully data-driven way.
Therefore, we use an architecture as shown in Fig.~\ref{fig:online_architecture}, where the current policy is controlling the car at 100Hz. At the same time we record the driving data and add it to a replay buffer which we use for off-policy updates using our \emph{policy output regularized}-SAC method. We start with an empty replay buffer and collect data with our sim-to-real policy for roughly 70 laps (or one full battery charge).
SAC, augmented with the policy output regularizer, is running as a secondary task on the same computer, updating its policy and (action) value function based on the replay buffer with 0.4 Hz. After 100 update steps of SAC, the controller policy is updated. Thus, we use the off-policy capability of SAC to have an approach that constantly updates the policy, while generating data that is close to on-policy which facilitates learning. Both the control loop and the SAC loop run on a laptop with a low power Intel i7-10510U CPU. On this hardware the SAC loop runs at roughly 40 Hz, and we update the policy about four times a lap.

\subsection{Policy Refinement Experimental Results}
%In this section we show the result of refining online the policy learned in simulation. 
We perform the experimental policy refinement as explained in Section \ref{sec:refinement}, and show that this approach indeed improves the autonomous racing policy, by updating the it purely data-driven. More precisely, we interact with the real race car for three hours, which is only 6\% of our simulation training. We warm-start with the sim-to-real policy that can already drive and is less damaging to the robotic platform than an random policy. Three hours of driving corresponds roughly to one million time steps and 450,000 SAC updates.

\begin{figure}[h]
    \vspace{-1 mm}
    \centering
    \includegraphics[width=0.35\textwidth]{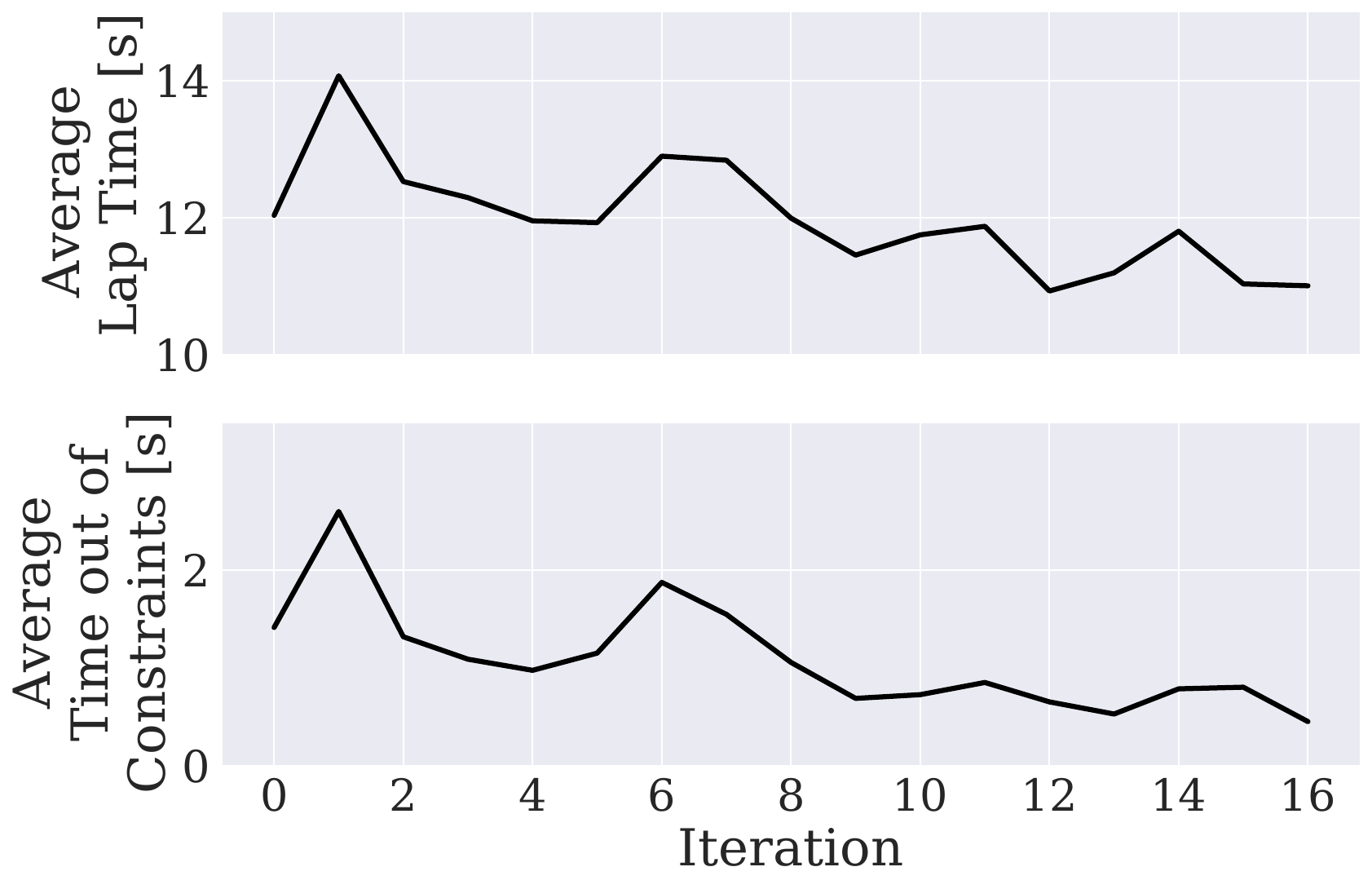}\\
    \caption{Progress of the performance during policy refinement. Each iteration is one averaged $\sim$12 min run, using one battery.}\label{fig:progress_online}
\end{figure}

\begin{figure}[h]
    \vspace{-1 mm}
    \centering
    \includegraphics[width=0.35\textwidth]{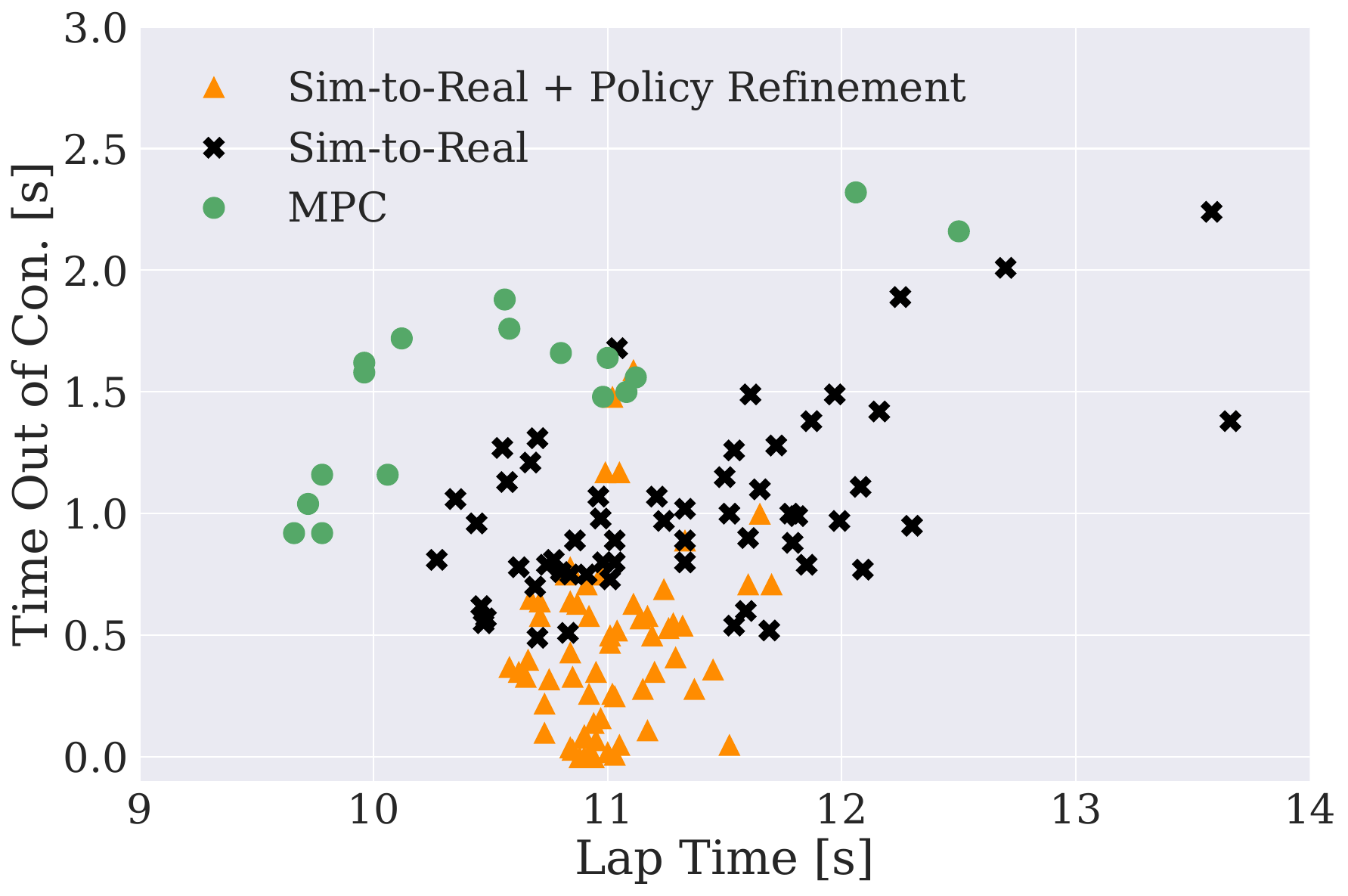}\\
    \caption{Performance comparison of the policy refined on the car.}\label{fig:performance_real}
\end{figure}

We show the learning performance in Fig.~\ref{fig:progress_online}, where each data-point is the average of the last 12 minutes of driving, which is also the interval at which we change the car batteries. The first and the last data points in Fig.~\ref{fig:progress_online} are for reference and show the initial sim-to-real policy and the final refined policy. In Fig.~\ref{fig:progress_online} we can clearly see that at the beginning of the learning phase the performance deteriorates, but the RL agent can recover and by the end of the three-hour interaction with the real car, the policy is drastically improved. Fig.~\ref{fig:performance_real} shows that both the lap time and the constraints violation benefit from the policy refinement: the RL policy shines at achieving very regular lap times and low number of constraint violations. Quantitatively, the refined policy achieved an average lap time of 11.01~s, and violated the constraints for only 0.44~s per lap. When we compare these numbers to the sim-to-real results, we see that the policy refinement reduced the mean lap time by 0.42~s and the constraint violations by 0.63~s. This is a drastic improvement, bringing the refined policy close to the MPC method in terms of lap time, while drastically reducing the number of constraint violations by 71\% compared the the state of the art MPC. It is especially interesting that the RL agent converged at a different type of solution than the hand-tuned MPC, emphasizing consistency and low constraint violations over raw peak lap times.

\vspace{-5 mm}

\section{Conclusion}\label{sec:conclusion}

%===============================================================================

In this work we successfully applied the RL algorithm SAC to the autonomous
racing problem. All results were validated experimentally on a miniature race track. We were able to achieve sim-to-real
through model randomization and a policy output regularization strategy.
Model randomization improves the generalization capability of the learned policy
while policy output regularization helps smoothing the actions without affecting
the convergence of the algorithm.
We also show that it is possible to improve the performance of the policy
trained in simulation by applying the same algorithm to the real car while it
drives around the track. For this purpose an asynchronous strategy is adopted
for storing and sampling driven trajectories to use for training.
The achieved performance on the platform is comparable to that achieved
by a state of the art model-based controller in terms of lap time, and shows an over three-fold improvement with respect to track constraints violations.

%===============================================================================

\bibliography{rl_racing}  % .bib
\bibliographystyle{IEEEtran}

\end{document}